*Original Article*

# Comparative Analysis of Non-Blind Deblurring Methods for Noisy Blurred Images

Poorna Banerjee Dasgupta

*Researcher, M.Tech Computer Science and Engineering, Nirma Institute of Technology, Gujarat, India*

***Abstract*** *- Image blurring refers to the degradation of an image wherein the image's overall sharpness decreases. Image blurring is caused by several factors. Additionally, during the image acquisition process, noise may get added to the image. Such a noisy and blurred image can be represented as the image resulting from the convolution of the original image with the associated point spread function, along with additive noise. However, the blurred image often contains inadequate information to uniquely determine the plausible original image. Based on the availability of blurring information, image deblurring methods can be classified as blind and non-blind. In non-blind image deblurring, some prior information is known regarding the corresponding point spread function and the added noise. The objective of this study is to determine the effectiveness of non-blind image deblurring methods with respect to the identification and elimination of noise present in blurred images. In this study, three non-blind image deblurring methods, namely Wiener deconvolution, Lucy-Richardson deconvolution, and regularized deconvolution were comparatively analyzed for noisy images featuring salt-and-pepper noise. Two types of blurring effects were simulated, namely motion blurring and Gaussian blurring. The said three non-blind deblurring methods were applied under two scenarios: direct deblurring of noisy blurred images and deblurring of images after denoising through the application of the adaptive median filter. The obtained results were then compared for each scenario to determine the best approach for deblurring noisy images.*

## I. INTRODUCTION

Image blurring involves image degradation wherein the overall sharpness of the image diminishes. Image blurring can be caused by several factors such as movement during the image capturing process (either by the camera or the target object), out-of-focus optics, atmospheric turbulence, and incorrect depth of field. In confocal microscopy, image blurring may occur owing to scattered light distortion [1–3]. Furthermore, during the image acquisition process, some noise may get added to the image. Such a blurred and noisy image can be represented by the following equation [4]:

$$B = I*p + n$$

Here, I represents the original image, B represents the blurred image, * indicates convolution, *p* denotes the associated point spread function (PSF), and *n* denotes the additive noise introduced during image acquisition. The PSF describes the response of an imaging system to a point source or object [5].

Generally, the blurred image contains insufficient information to uniquely determine the original image, thereby making it an ill-posed problem [6]. Depending on the availability of blurring information, image deblurring methods can be categorized as blind and non-blind. In non-blind deblurring, some prior information is known regarding the corresponding PSF and the additive noise. In contrast, blind image deblurring involves recovering the sharp original image from the blurred and noisy image without any prior knowledge regarding the PSF or the additive noise [7]. Furthermore, image deblurring is often an iterative process. The deblurring process may have to be repeated multiple times to vary the parameters specified to the deblurring method with each iteration, until an image is achieved that is the closest approximation of the original image.

It should be noted that during the deblurring process, new features and artifacts (such as the ringing effect and checkered effect) may appear in the deblurred image that were not present in the original image. The ringing effect can often appear in deblurred images, and it occurs when the deblurring method involves the discrete Fourier transform during deconvolution, which causes truncation of certain high frequencies near the edges present in an image, thereby leading to the appearance of ripples or "ringing." Similarly, the checkered effect or pattern is caused by pixel replication during deconvolution [8].

In this study, three non-blind image deblurring methods, namely Wiener deconvolution, Lucy-Richardson deconvolution, and regularized deconvolution were comparatively analyzed for noisy blurred images. The said three methods, which are described in further detail in Section II, were applied for deblurring under two scenarios: direct deblurring of the noisy blurred images and deblurring after denoising the blurred images. The objective of doing thus is to determine the effectiveness of non-blind image deblurring methods with respect to the identification and elimination of noise present in blurred images. Two types of blurring effects were simulated, namely motion blurring and Gaussian blurring. Furthermore, salt-and-pepper noise was added to the blurred images. Salt-and-pepper noise, also known as impulse noise, is caused by sharp and sudden disturbances in the image signal. It presents itself as sparsely occurring black and white pixels in the image and can be difficult to eliminate during image deblurring [9, 10]. In this study, the adaptive median filter was employed for denoising the blurred images featuring salt-and-pepper noise. Adaptive median filtering is a spatial non-linear digital filtering technique wherein the neighborhood size of the filter kernel can be varied depending on the extent of the desired image smoothness; larger the neighborhood size, greater or more severe will be the smoothing effect.

In Section II, the aforementioned three non-blind image deblurring methods are described, along with their respective application to motion-blurred and Gaussian-blurred images containing salt-and-pepper noise. Furthermore, their effectiveness (in terms of complete image restoration) is examined under two scenarios. Section III comprises an in-depth analysis of the obtained results. Finally, in Section IV, the conclusions derived from the proposed study are elucidated, and the future scope of work is discussed.

## II. NON-BLIND IMAGE DEBLURRING

In this study, a sample grayscale image of a flower was incorporated for applying various deblurring techniques through MATLAB programming and simulation. Fig. 1 shows the original grayscale image. Fig. 2(a) illustrates the effect of motion blurring and Fig. 2(b) illustrates the effect of adding salt-and-pepper noise to the motion-blurred image. Similarly, Fig. 3(a) depicts the effect of Gaussian blurring and Fig. 3(b) illustrates the effect of adding salt-and-pepper noise to the Gaussian-blurred image. While simulating motion blurring due to camera movement, a linear displacement of 15 px was applied to the original image, along with an angular displacement of 11° in the counter-clockwise direction, resulting in a motion blurring PSF filter of size 5×15. Furthermore, for simulating blurring caused by out-of-focus optics, a Gaussian low-pass PSF filter of size 5×5 with a standard deviation of 7 was applied. Subsequently, salt-and-pepper noise with noise density of 7% was added to both the motion-blurred and Gaussian-blurred images. Table 1 summarizes various statistical attributes associated with these images.

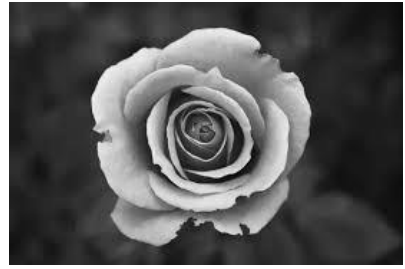

**Fig. 1 Original grayscale image**

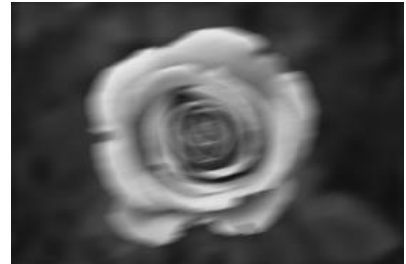

**Fig. 2(a) Effect of motion blurring**

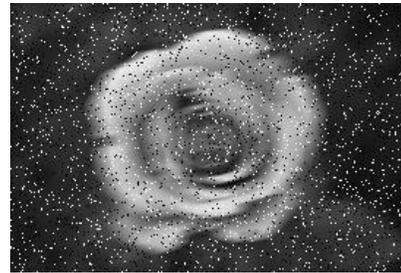

**Fig. 2(b) Addition of salt-and-pepper noise after motion blurring**

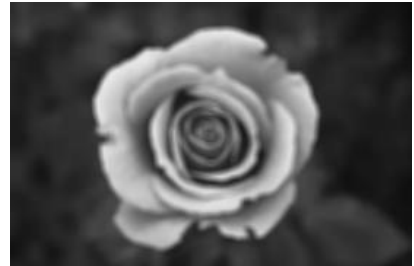

**Fig. 3(a) Effect of Gaussian blurring**

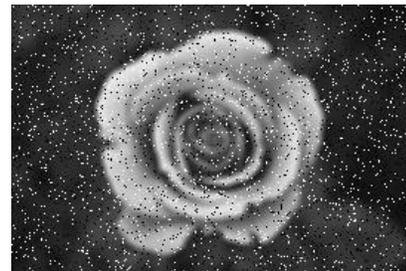

**Fig. 3(b) Addition of salt-and-pepper noise after Gaussian blurring**

**Table 1. Statistical attributes of original, blurred, and noisy images**

| Image | Attributes |
|---|---|
| **Original Image** (Fig. 1) | Image Dimensions: 183×275<br>Mean Value (μ): 79.5248<br>Standard Deviation (σ): 62.0037<br>Signal-to-noise Ratio (μ/σ): 1.2826 |
| **Motion-Blurred Image** (Fig. 2(a)) | Linear displacement of 15 px and angular displacement of 11° in counter-clockwise direction was applied.<br>Image Dimensions: 183×275<br>Mean Value (μ): 79.5397<br>Standard Deviation (σ): 55.6381<br>Signal-to-noise Ratio (μ/σ): 1.4296 |
| **Noisy Motion-Blurred Image** (Fig. 2(b)) | Salt-and-pepper noise with 7% noise density was added.<br>Image Dimensions: 183×275<br>Mean Value (μ): 82.8611<br>Standard Deviation (σ): 64.7236<br>Signal-to-noise Ratio (μ/σ): 1.2802 |
| **Gaussian-Blurred Image** (Fig. 3(a)) | A Gaussian low-pass filter of size 5×5 with a standard deviation of 7 was applied.<br>Image Dimensions: 183×275<br>Mean Value (μ): 79.2691<br>Standard Deviation (σ): 57.7991<br>Signal-to-noise Ratio (μ/σ): 1.3715 |
| **Noisy Gaussian-Blurred Image** (Fig. 3(b)) | Salt-and-pepper noise with 7% noise density was added.<br>Image Dimensions: 183×275<br>Mean Value (μ): 82.9601<br>Standard Deviation (σ): 66.8557<br>Signal-to-noise Ratio (μ/σ): 1.2409 |

As described earlier, to analyze the said three non-blind image deblurring methods, each method was applied to both the noisy and the denoised versions of the blurred images. Furthermore, to eliminate the added salt-and-pepper noise from the blurred images, the adaptive median filter was incorporated, whose kernel size was varied between 3×3 to 5×5. Fig. 4(a) and 4(b) depict the denoised motion-blurred and Gaussian-blurred images after the application of the adaptive median filter, respectively. To indicate how effectively the adaptive median filter can eliminate salt-and-pepper noise, Table 2 summarizes the statistical attributes of the denoised images in comparison with that of the corresponding noise-free blurred images, where the root mean squared error (RMSE) values represent the mean difference between the pixel intensity values of the images.

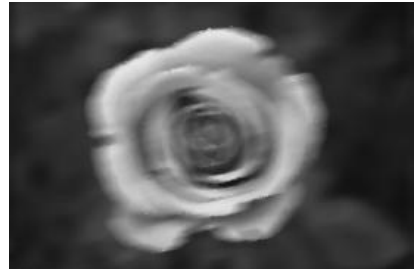

**Fig. 4(a) Denoised motion-blurred image**

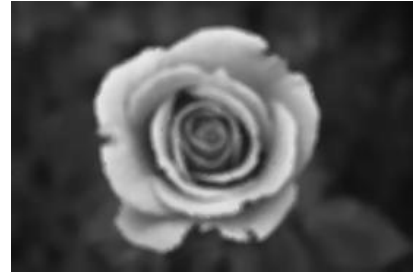

**Fig. 4(b) Denoised Gaussian-blurred image**

**Table 2. Statistical attributes of blurred and denoised images**

| | Noise-Free Blurred Image | Denoised Blurred Image |
|---|---|---|
| **Motion Blurring** | Image Dimensions: 183×275<br><br>Mean Value (μ): 79.5397<br><br>Standard Deviation (σ): 55.6381<br><br>Signal-to-noise Ratio (μ/σ): 1.4296 | Image Dimensions: 183×275<br><br>Mean Value (μ): 79.5264<br><br>Standard Deviation (σ): 55.5289<br><br>Signal-to-noise Ratio (μ/σ): 1.4321 |
| | **Root Mean Squared Error (RMSE) = 1.9863** | |
| **Gaussian Blurring** | Image Dimensions: 183×275<br><br>Mean Value (μ): 79.2691<br><br>Standard Deviation (σ): 57.7991<br><br>Signal-to-noise Ratio (μ/σ): 1.3715 | Image Dimensions: 183×275<br><br>Mean Value (μ): 79.3717<br><br>Standard Deviation (σ): 57.8279<br><br>Signal-to-noise Ratio (μ/σ): 1.3725 |
| | **Root Mean Squared Error (RMSE) = 1.8751** | |

*A. Wiener Deconvolution*

Wiener deconvolution (named after the mathematician and philosopher Norbert Wiener) is a classic non-blind image deblurring method [11]. Wiener deconvolution can be employed for producing an estimate or approximation of the desired original image through linear time-invariant filtering of the observed noisy blurred image, given that the corresponding PSF and the additive noise are known to some extent. Wiener deconvolution is based on the minimization of the mean squared error between the estimated image and the original image.

In this study, to analyze the effectiveness of Wiener deconvolution in terms of image deblurring of noisy blurred images, Wiener deconvolution was applied under two scenarios. In the first case, Wiener deconvolution was directly applied to the noisy motion-blurred image (i.e., Fig. 2(b)) and the noisy Gaussian-blurred image (i.e., Fig. 3(b)).

In the second case, Wiener deconvolution was applied to the denoised motion-blurred image (i.e., Fig. 4(a)) and the denoised Gaussian-blurred image (i.e., Fig. 4(b)). Fig. 5(a) and 5(b) depict the output images after direct application of Wiener deconvolution to the noisy motion-blurred image and noisy Gaussian-blurred image, respectively. Fig. 5(c) and 5(d) indicate the output images after applying Wiener deconvolution to the denoised motion-blurred and Gaussian-blurred images, respectively. Furthermore, Table 3 lists the statistical attributes associated with the output images of Wiener deconvolution.

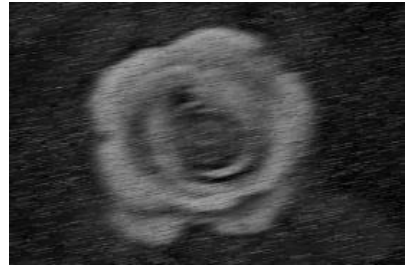

**Fig. 5(a) Wiener deconvolution applied to noisy motion-blurred image**

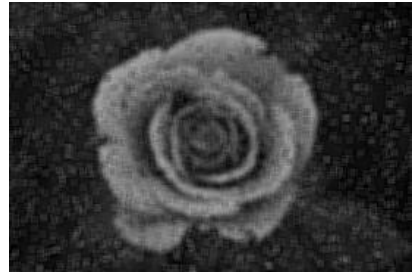

**Fig. 5(b) Wiener deconvolution applied to noisy Gaussian-blurred image**

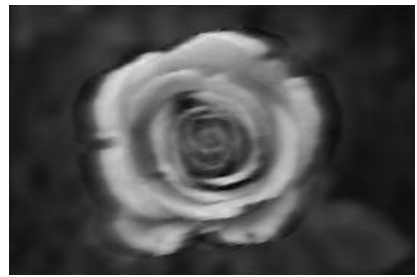

**Fig. 5(c) Wiener deconvolution applied to denoised motion-blurred image**

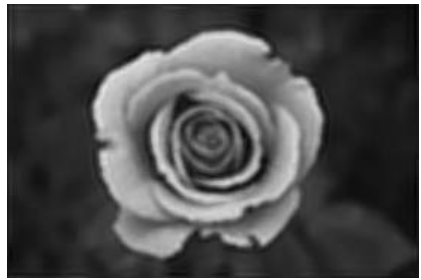

**Fig. 5(d) Wiener deconvolution applied to denoised Gaussian-blurred image**

**Table 3. Statistical attributes of output images of Wiener deconvolution**

|  | **Image Dimensions** | **Mean Value (μ)** | **Standard Deviation (σ)** | **Signal-to-noise Ratio (μ/σ)** |
|---|---|---|---|---|
| **Fig. 5(a)** | 183×275 | 54.7351 | 35.8125 | 1.5284 |
| **Fig. 5(b)** | 183×275 | 54.8173 | 37.2535 | 1.4715 |
| **Fig. 5(c)** | 183×275 | 72.2956 | 52.3488 | 1.3811 |
| **Fig. 5(d)** | 183×275 | 72.2666 | 54.2785 | 1.3314 |

### *B. Lucy-Richardson Deconvolution*

Lucy-Richardson deconvolution is an iterative procedure for recovering an underlying image that has been blurred by a known PSF (along with known additive noise). It was named after William Richardson and Leon Lucy [12, 13]. The algorithm of Lucy-Richardson deconvolution maximizes the likelihood that the blurred image when convolved with the PSF, is an instance of the original image under Poisson statistics. However, noise amplification is a common issue associated with maximum likelihood methods that attempt to fit data as closely as possible. After several iterations, the restored image may have a speckled appearance, which is caused by fitting the noise present in the image too closely. The ringing effect may also appear.

In this study, to examine the effectiveness of Lucy-Richardson deconvolution in terms of image restoration (deblurring and denoising), Lucy-Richardson deconvolution was applied under two scenarios. In the first case, Lucy-Richardson deconvolution was directly applied to the noisy motion-blurred image (i.e., Fig. 2(b)) and the noisy Gaussian-blurred image (i.e., Fig. 3(b)). In the second case, Lucy-Richardson deconvolution was applied to the denoised motion-blurred image (i.e., Fig. 4(a)) and the denoised Gaussian-blurred image (i.e., Fig. 4(b)). Fig. 6(a) and 6(b) depict the output images after direct application of Lucy-Richardson deconvolution to the noisy motion-blurred image and noisy Gaussian-blurred image, respectively. Fig. 6(c) and 6(d) indicate the output images after applying Lucy-Richardson deconvolution to the denoised motion-blurred and Gaussian-blurred images, respectively.

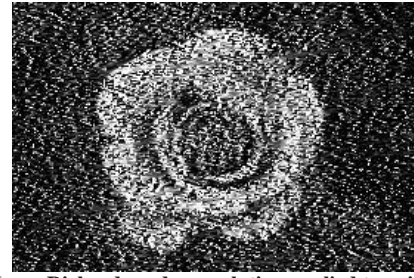

**Fig. 6(a) Lucy-Richardson deconvolution applied to noisy motion-blurred image**

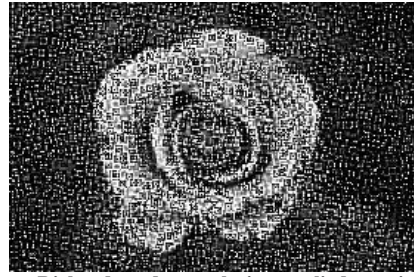

**Fig. 6(b) Lucy-Richardson deconvolution applied to noisy Gaussian-blurred image**

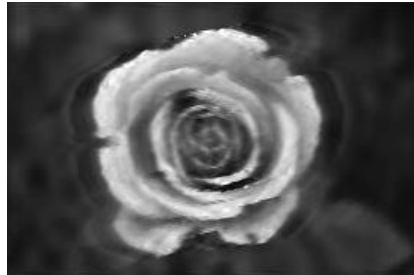

**Fig. 6(c) Lucy-Richardson deconvolution applied to denoised motion-blurred image**

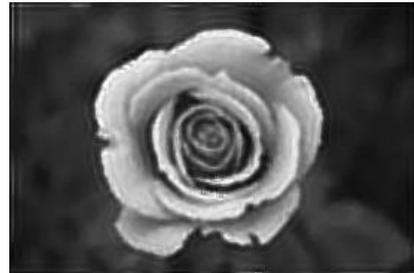

**Fig. 6(d) Lucy-Richardson deconvolution applied to denoised Gaussian-blurred image**

Table 4 lists the statistical attributes associated with the output images of Lucy-Richardson deconvolution. It should be noted that while applying Lucy-Richardson deconvolution directly to the noisy blurred images, the best results were obtained after fifteen iterations. However, while applying Lucy-Richardson deconvolution to the denoised blurred images, the best results were obtained after ten iterations. The same has been indicated in Table 4.

**Table 4. Statistical attributes of output images of Lucy-Richardson deconvolution**

| | Image Dimensions | Mean Value (μ) | Standard Deviation (σ) | Signal-to-noise Ratio (μ/σ) |
|---|---|---|---|---|
| **Fig. 6(a)** (No. of iterations = 15) | 183×275 | 74.3681 | 83.3266 | 0.8925 |
| **Fig. 6(b)** (No. of iterations = 15) | 183×275 | 74.0429 | 84.2178 | 0.8792 |
| **Fig. 6(c)** (No. of iterations = 10) | 183×275 | 79.5036 | 59.4781 | 1.3367 |
| **Fig. 6(d)** (No. of iterations = 10) | 183×275 | 79.5258 | 60.8151 | 1.3077 |

*C. Regularized Deconvolution*

As mentioned earlier in Section I, the process of image deblurring via deconvolution for obtaining the original image from the noisy blurred image is an ill-posed problem. To eliminate implausible solutions and to help the iterative deconvolution process converge, regularization is often required. Regularization can help in reducing ringing and noise amplification, which are often caused during iterative deconvolution. Regularization can be conducted in several ways, ranging from the incorporation of image priors to using weighting factors corresponding to image smoothness constraints [14–16]. Furthermore, regularized deconvolution can be helpful when limited information is known about the additive noise and corresponding PSF.

In this study, to conduct regularized deconvolution, the concepts of Lagrange multipliers and Laplacian regularization operator were incorporated. Since the objective of the deblurring process is to find the optimal or closest approximation of the original image from the blurred image, the method of Lagrange multipliers can serve as a strategy for finding the local maxima and minima of the blurring PSF when subject to image smoothness constraints [17]. In general, to find the maximum or minimum of a function $f(x)$ subjected to the equality constraint $g(x)=0$, the Lagrangian function ($L$) can be defined as follows [18]:

$$L(x, \lambda) = f(x) - \lambda g(x)$$

Here, $\lambda$ represents the Lagrange multiplier. Furthermore, Laplacian regularization, which is based on the finite-difference approximation of the Laplacian operator, can be helpful in constraining the least squares optimization associated with image deblurring [19, 20].

To analyze the effectiveness of regularized deconvolution, it was applied under two scenarios. In the first case, regularized deconvolution was directly applied to the noisy motion-blurred image (i.e., Fig. 2(b)) and the noisy Gaussian-blurred image (i.e., Fig. 3(b)). In the second case, regularized deconvolution was applied to the denoised motion-blurred image (i.e., Fig. 4(a)) and the denoised Gaussian-blurred image (i.e., Fig. 4(b)). Fig. 7(a) and 7(b) depict the output images after direct application of regularized deconvolution to the noisy motion-blurred image and noisy Gaussian-blurred image, respectively. Fig. 7(c) and 7(d) indicate the output images after applying regularized deconvolution to the denoised motion-blurred and Gaussian-blurred images, respectively. Furthermore, Table 5 lists the statistical attributes associated with the output images of regularized deconvolution, wherein the estimated Lagrange multiplier value ($\lambda$) corresponding to each image has also been specified. While implementing regularized deconvolution, the dimensions of the Laplacian regularization operator kernel matched that of the non-singleton dimensions of the corresponding PSF. Hence, in the case of motion blurring, the size of the Laplacian regularization operator kernel was 5×15, whereas its size was 5×5 in the case of Gaussian blurring. The same has been specified in Table 5 as well.

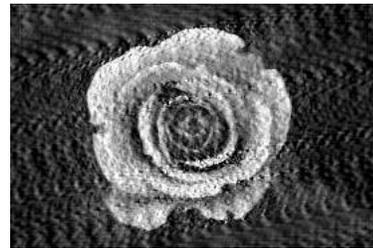

**Fig. 7(a) Regularized deconvolution applied to noisy motion-blurred image**

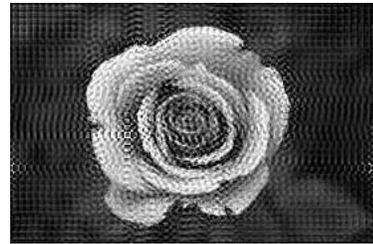

**Fig. 7(b) Regularized deconvolution applied to noisy Gaussian-blurred image**

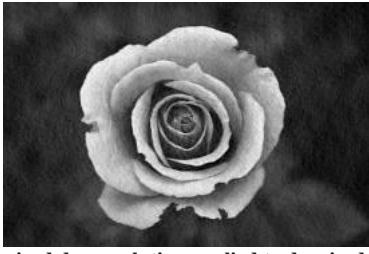

**Fig. 7(c) Regularized deconvolution applied to denoised motion-blurred image**

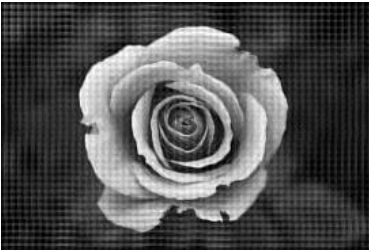

**Fig. 7(d) Regularized deconvolution applied to denoised Gaussian-blurred image**

**Table 5. Statistical attributes of output images of regularized deconvolution**

|  | Fig. 7(a) | Fig. 7(b) | Fig. 7(c) | Fig. 7(d) |
|---|---|---|---|---|
| **Image Dimensions** | 183×275 | 183×275 | 183×275 | 183×275 |
| **Laplacian Regularization Operator Kernel Dimensions** | 5×15 | 5×5 | 5×15 | 5×5 |
| **Lagrange Multiplier ($\lambda$)** | 2.56e-04 | 1.44e-04 | 3.49e-05 | 3.28e-05 |
| **Mean Value ($\mu$)** | 83.3865 | 82.8682 | 79.6606 | 79.2255 |
| **Standard Deviation ($\sigma$)** | 60.4309 | 60.3491 | 65.8695 | 63.4596 |
| **Signal-to-noise Ratio ($\mu/\sigma$)** | 1.3799 | 1.3731 | 1.2094 | 1.2484 |

### III. RESULT ANALYSIS

- It can be observed from Fig. 4(a), 4(b), and Table 2 that the denoising of the motion-blurred and Gaussian-blurred images was effectively conducted by employing the adaptive median filter. Based on the results listed in Table 2, it is observed that the values of the statistical attributes (image mean pixel value, standard deviation, and signal-to-noise ratio) of the denoised images are very close to the corresponding values of the noise-free blurred image counterparts (i.e., Fig. 2(a) and 3(a)). This observation is further supported by the low RMSE values of difference in pixel intensity values between the denoised and noise-free blurred images, as is listed in Table 2.

- *Wiener deconvolution*: It is observed that the images obtained by direct application of Wiener deconvolution (i.e., Fig. 5(a) and 5(b)) are visibly darker than the original grayscale image (Fig. 1), as is indicated by significantly lower mean pixel intensity values listed in Table 3. Both Fig. 5(a) and Fig. 5(b) have a spotted appearance, indicating that the added salt-and-pepper noise could not be completely eliminated. On the other hand, the images (i.e., Fig. 5(c) and 5(d)) obtained by applying Wiener deconvolution to the denoised blurred images do not have a spotted appearance. However, these images appear to be partially blurred and exhibit the ringing effect near the flower's petal boundaries, which can be prominently noticed in Fig. 5(c) compared to Fig. 5(d).

- *Lucy-Richardson deconvolution*: It is observed that the images obtained by direct application of Lucy-Richardson deconvolution (i.e., Fig. 6(a) and 6(b)) have a speckled appearance, which is caused by fitting the noise present in the images very closely. The images obtained (i.e., Fig. 6(c) and 6(d)) by applying Lucy-Richardson deconvolution to the denoised blurred images do not have a speckled appearance, but they exhibit the ringing effect near the flower's petal boundaries, which can be seen more prominently in Fig. 6(c) as compared to Fig. 6(d). Furthermore, based on the results listed in Table 4, it is observed that fewer iterations are required for processing denoised blurred images (ten iterations) than for noisy blurred images (fifteen iterations).

- *Regularized deconvolution*: It is observed that the images obtained by direct application of regularized deconvolution (i.e., Fig. 7(a) and 7(b)) have repetitive white artifacts throughout both the images that were not present in the original grayscale image, thereby causing higher mean pixel intensity values (as listed in Table 5). The images obtained (i.e., Fig. 7(c) and 7(d)) by applying regularized deconvolution to the denoised blurred images do not feature such white artifacts. However, a prominent checkerboard pattern can be seen in Fig. 7(d), which is caused by pixel replication during deconvolution. Furthermore, based on the results listed in Table 5, it is observed that the estimated values of the Lagrange multiplier are lower while processing the denoised blurred images than those while processing the noisy blurred images, which implies smaller error or difference between the pixel intensity values of the original and denoised deblurred images.

## IV. CONCLUSION

In this study, three non-blind image deblurring methods, namely Wiener deconvolution, Lucy-Richardson deconvolution, and regularized deconvolution were analyzed for noisy blurred images. Based on the obtained results, it is concluded that all three methods perform poorly when salt-and-pepper noise is present in the blurred images, even though the associated PSFs and additive noise characteristics are known beforehand. However, the qualitative results of image deblurring improve significantly when the said methods are applied to denoised blurred images. Both Wiener deconvolution and Lucy-Richardson deconvolution cause more prominent ringing while deblurring motion-blurred images as compared to deblurring Gaussian-blurred images. Additionally, in the case of Lucy-Richardson deconvolution, fewer iterations are required for processing denoised blurred images than for noisy blurred images. In terms of visual quality and image restoration, the best results are obtained upon application of regularized deconvolution to denoised motion-blurred images. As future work, the said non-blind deblurring methods can be comparatively analyzed for colored images featuring various types of additive noises and blurring effects. Furthermore, different approaches for blind image deblurring can be compared to determine which approach yields the best results.